\documentclass[a4paper,fleqn]{cas-dc}
\usepackage[numbers]{natbib}
\usepackage{booktabs}
\usepackage{multirow}
\usepackage{float}
\usepackage{mathtools}
\usepackage{soul}
\usepackage{color}
\usepackage{enumitem}
\usepackage{hyperref}
\usepackage{diagbox}
\usepackage{makecell}
\usepackage[toc,page]{appendix}
\usepackage[colorinlistoftodos]{todonotes}
\usepackage{subfigure}

\usepackage{algorithm} 
\usepackage[algo2e]{algorithm2e} 

\usepackage{algpseudocode}
\usepackage{amsmath}

\usepackage{amssymb}
\usepackage{mathtools}
\usepackage[switch]{lineno}
\usepackage{mathrsfs}
\usepackage{bm}
\usepackage{lineno}
\usepackage{scrtime}
\usepackage{etoolbox}

\begin{document}
\let\WriteBookmarks\relax
\def\floatpagepagefraction{1}
\def\textpagefraction{.001}
\shorttitle{ComSimGNN}
\shortauthors{Haoyan Xu and Runjian Chen et~al.}
\title [mode = title]{CoSimGNN: Towards Large-scale Graph Similarity Computation}       
\address[1]{School of Big Data and Software Engineering, Chongqing University, Chongqing, 401331, China}
\address[2]{College of Energy Engineering, Zhejiang University, Zhejiang, 310027, China}

\cortext[cor1]{These authors contributed equally to this research.}
\cortext[cor1]{Corresponding author. E-mail address: yueyangw@cqu.edu.cn}

\author[1,2]{Haoyan Xu}
\cormark[1]

\author[1,2]{Runjian Chen}
\cormark[1]

\author[1]{Yueyang Wang}[orcid=0000-0003-3210-0930]
\cormark[2]

\author[1,2]{Ziheng Duan}

\author[1,2]{Jie Feng}

\begin{abstract}
The ability to compute similarity scores between graphs based on metrics such as Graph Edit Distance (GED) is important in many real-world applications. Computing exact GED values is typically an NP-hard problem and traditional algorithms usually achieve an unsatisfactory trade-off between accuracy and efficiency. Recently, Graph Neural Networks (GNNs) provide a data-driven solution for this task, which is more efficient while maintaining prediction accuracy in small graph (around 10 nodes per graph) similarity computation. Existing GNN-based methods, which either respectively embed two graphs (lack of low-level cross-graph interactions) or deploy cross-graph interactions for whole graph pairs (redundant and time-consuming), are still not able to achieve competitive results when the number of nodes in graphs increases. In this paper, we focus on similarity computation for large-scale graphs and propose the "embedding-coarsening-matching" framework \textsc{CoSimGNN}, which first embeds and coarsens large graphs with adaptive pooling operation and then deploys fine-grained interactions on the coarsened graphs for final similarity scores. Furthermore, we create several synthetic datasets which provide new benchmarks for graph similarity computation. Detailed experiments on both synthetic and real-world datasets have been conducted and \textsc{CoSimGNN} achieves the best performance while the inference time is at most \textbf{1/3} of that of previous state-of-the-art.
\end{abstract}


\maketitle

\section{Introduction}
With flexible representative abilities, graphs have a broad range of applications in various fields \cite{emmert2016fifty}, such as social network study, computational chemistry \cite{gilmer2017neural}, and biomedical image analysis \cite{ktena2017distance}.
Graph similarity computation, i.e. the problem to predict similarity
between arbitrary pairs of graph-structured objects, is one of the fundamental challenges and appears in various real-world applications, including biological molecular similarity search \cite{kriegel2004similarity,tian2007saga} and social group network similarity identification \cite{steinhaeuser2008community, ogaard2013discovering,chen2015kappa}.

To compute graph-graph similarity, evaluation metrics such as Graph Edit Distance (GED) \cite{6313167} were developed. Traditional methods for these metrics can be divided into two classes.
The first class calculates the exact values\cite{riesen2013novel,mccreesh2017partitioning}. 
While the exact similarity scores for sure help us better understand the relationship between graphs, the problem of exponential time complexity of exact graph similarity computation remains. 
The second class only computes the approximate values and saves time in return\cite{10.1007/11815921_17,jonker1987shortest,10.1007/978-3-642-20844-7_11,kuhn1955hungarian,riesen2009approximate}.
However, these algorithms still run with polynomial or even sub-exponential time complexity.

\begin{figure}
\centering
\includegraphics[width=1\linewidth]{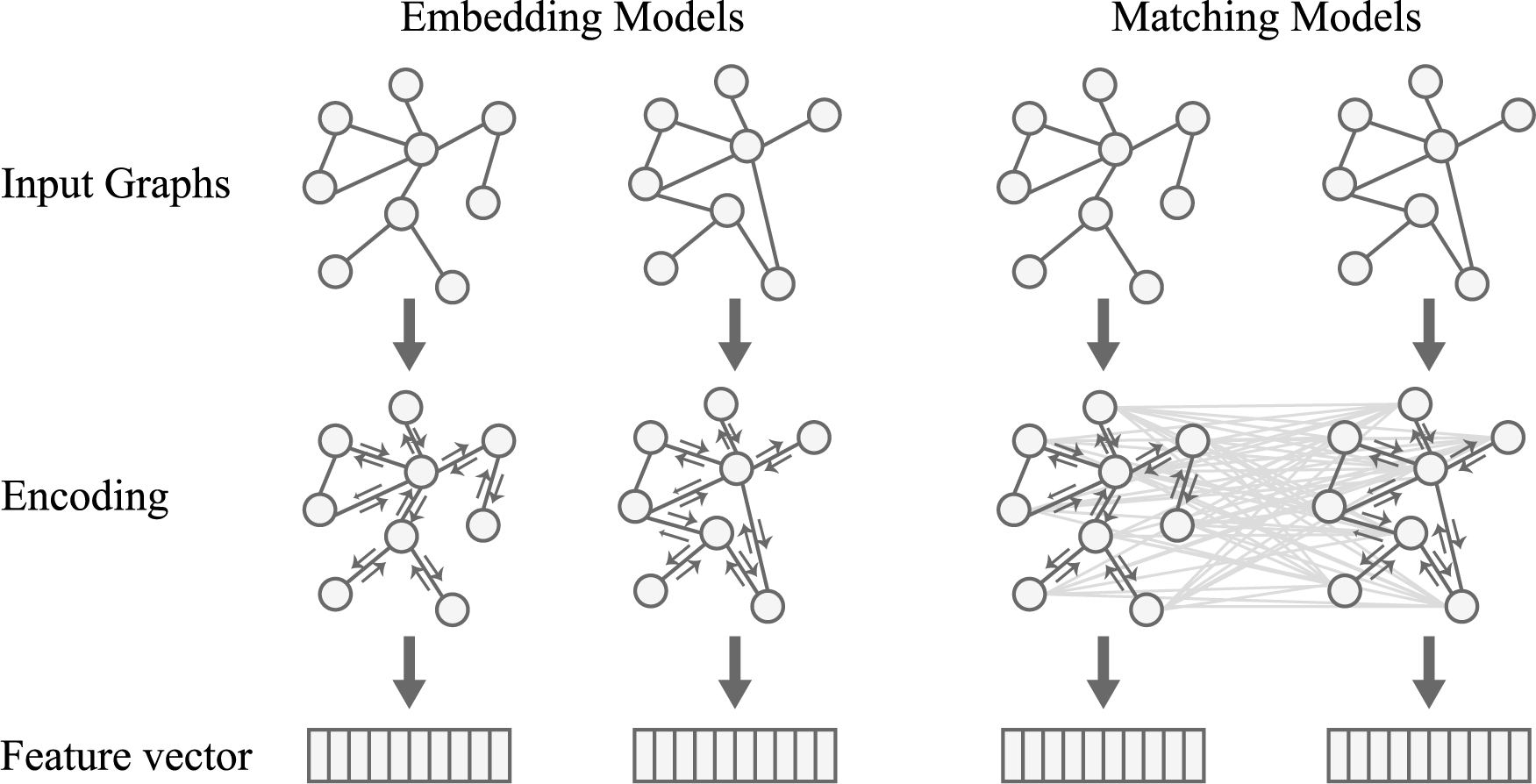}
\caption{Two categories of existing data-driven models for graph similarity computation.}
\label{fig:aec}
\end{figure}

With the rapid development of deep learning technology, graph neural networks (GNNs) which automatically extract the graph’s structural characteristics provide a new solution for similarity computation of graph structures. Typically, GNN-based solution involves two stages: (1) embedding, which maps each graph to its representation feature vector, making similar graphs close in the vector space; (2) similarity computation based on these feature vectors. With different methods in the first stage, existing deep learning techniques for graph similarity computation can be classified into two categories, as illustrated in Fig. \ref{fig:aec}.
The first category (named embedding models) exemplified by Hierarchically Mean (\textsc{GCN-Mean}) and Hierarchical Max (\textsc{GCN-Max}) \cite{NIPS2016_6081}, directly maps graphs to feature vectors respectively by hierarchically coarsening the graphs and computes similarity based on the representation vectors of the two graphs.
The second category (named matching models) embraced by Graph Matching Networks (\textsc{GMN}) \cite{li2019graph} embeds pair of graphs at the same time with cross-graph low-level interactions.
Other methods like \textsc{GSimCNN} \cite{bai2018convolutional} and \textsc{GRAPHSIM} \cite{bai2020learning} are similar to the second category except that they do not generate two separate feature vectors but instead directly embed graph pair to similarity score with cross-graph interactions.
Meanwhile \textsc{SimGNN} \cite{bai2018simgnn} deploy the two categories above parallelly.

In general, the embedding models are efficient but not effective, 
and the matching models are generally more effective but not efficient. 
For example, we assume that there are $K$ graphs in the database. The graph similarity search problem is to find out the most similar graph with the new graph by computing the similarity score between this new graph and all $K$ graphs.
For embedding models, they can embed all the graphs in the database to their own feature vectors in advance without considering cross-graph fine-grained interactions. So that they only need to forward the new graph to its feature vector and compute similarity $K$ times based on these feature vectors but reducing the accuracy.
Meanwhile, matching models have to forward the new graph respectively with all graphs in the database (cross-graph interactions can not be finished in advance). So that the performance of matching models is generally higher than that of embedding models but they are time-consuming. 

In real world, large-scale graphs (graph with hundreds of nodes) similarity computation is common and significant in domains such as bio-informatics, pattern recognition, social network, etc. However, existing methods \cite{bai2018simgnn, bai2020learning,bai2018convolutional,li2019graph} conduct their experiments of graph-graph similarity tasks mostly on relatively small graphs, where there are only around 10 nodes in each graph.
For large-scale graphs, it is difficult for embedding models to capture subtle structural difference between two graphs. And the learned embedding is not easy to express all the features, leading to a decline in accuracy. For matching models, they consume a lot of time due to the pairwise cross-graph interactions among nodes (at least quadratic time complexity over number of nodes). Hence, the first challenge for large-scale graph similarity computation is to balance the efficiency and effectiveness.

Different from small graphs, large-scale graph may contain different sub-graph structure, leading to diverse local features. Directly embedding the whole graph is not intuitively a proper way. Many existing graph pooling based methods \cite{Khasahmadi2020Memory-Based} have been proposed to aggregate the hierarchical features by involving centroids for cluster assignments. However, they never rely on the input graphs. This does not make sense intuitively because the centroids for different graphs should be different.
Thus, for large-scale graph similarity computation, it is challenging to capture fine-grained and hierarchical graph features and fuse them adaptively.

To this end, we focus on similarity computation problem on relatively large-scale graphs and propose our "embedding-coarsening-matching" graph similarity computation framework \textbf{\textsc{CoSim-GNN}}, i.e., \emph{\underline{Co}}arsening-based \emph{\underline{Sim}}ilarity Computation via \emph{\underline{G}}raph \emph{\underline{N}}eural \emph{\underline{N}}etworks. 
To better use the hierarchical nature of large-scale graphs, \textsc{CoSim-GNN} first introduces graph embedding layers and pooling layers with intra-attention mechanism to encode and coarsen the graphs. Specially, on the one hand, the pooling layers are designed to be adaptive, where the generation process of centroids is related on input graphs. On the other hand, the pooling layers guarantee that the centroids of the same graph would not change when the permutation of nodes changes (permutation invariance). Finally, we designed a matching mechanism on the coarsened graph pair to compute similarity between graphs with inter-attention mechanism.
We highlight our main contributions as follows:
\begin{itemize}[leftmargin=*]
\item We propose a novel framework, which first hierarchically encodes and coarsens graphs with adaptive pooling operation and then deploys a matching mechanism on the coarsened graph pairs, to address the challenging problem of similarity computation between large-scale graphs.
\item We create a set of synthetic datasets and collect a set of real-world datasets with different sizes for both graph-graph classification and similarity regression tasks which provides new benchmarks to study graph similarity computation.
\item We conduct extensive experiments on real-world datasets and synthetic datasets consisting of large graphs. Our framework shows significant improvement in time complexity as compared to matching models. It also outperforms matching models (thus far better than embedding models) on similarity regression and classification tasks.
\item Our framework is able to learn from graph pairs with a relatively small number of nodes (approximate 100) and then be deployed to infer similarity between huge graphs with thousands of nodes.
\end{itemize}

\section{Related Work}
\subsection{Graph Similarity Computation}
\subsubsection{Graph Similarity Metrics} Graph edit distance (GED) as a similarity measure is preferred over other distances or similarity measures because of its generality and broad applicability \cite{gouda2016csi_ged}. The Graph Edit Distance (GED) is defined as the minimum cost taken to transform one graph to the other via a sequence graph edit operations. Here an edit operation on a graph is an insertion or deletion of a vertex/edge or relabelling of a vertex. 
\subsubsection{Traditional Algorithms} Various traditional heuristic methods have been proposed for approximation of graph similarity, including \cite{10.1007/11815921_17}, \cite{jonker1987shortest}, \cite{10.1007/978-3-642-20844-7_11}, \cite{kuhn1955hungarian} and \cite{riesen2009approximate}. These algorithms aim to reduce the time complexity with the sacrifice in accuracy. However, they always achieve an unsatisfactory trade-off between accuracy and efficiency.
\subsubsection{Data-driven Methods}
Data-driven methods can be classified into embedding models and matching models, and they have the key advantage of efficiency due to the nature of neural network computation. However, existing methods are still not easily scalable to large-scale graphs. Embedding models such as \cite{NIPS2016_6081} are efficient but not effective due to the lack of cross-graph low-level interactions. Matching models such as \cite{li2019graph}, \cite{bai2018convolutional}, \cite{bai2020learning} and \cite{bai2018simgnn}, are generally effective but not efficient due to the need of low-level interactions across whole graphs. 

\subsection{Graph Pooling}
In order to make our method efficient and easily scalable to large-scale graphs, we adopt graph pooling layers to coarsen original graphs and then deploy fine-grained cross-graph interactions on these coarsened (smaller and denser) graphs. 
Graph pooling (downsampling) operations aim at reducing the number of graph nodes and learning new representations. 
\textsc{DiffPool} \cite{ying2018hierarchical} treats the graph pooling as a node clustering problem and trains two parallel GNNs to compute node representations and cluster assignments. \textsc{gPool} \cite{gao2019graph} and \textsc{SAGPool} \cite{lee2019self} drop nodes from the original graph rather than group multiple nodes to form a cluster in the pooled graph. They devise a top-K node selection procedure to form an induced sub-graph for the next input layer. Although they are more efficient than \textsc{DiffPool}, they do not aggregate nodes nor compute soft edge weights. This makes them unable to preserve node and edge information effectively. Based on \textsc{DiffPool}, \textsc{MemPool} \cite{Khasahmadi2020Memory-Based} also learns soft cluster assignments, and uses a clustering-friendly distribution to compute the attention scores between nodes and clusters. However, \textsc{MemPool} generates centroids (which are used to compute soft assignments) without involving the input graphs, which does not make sense intuitively because the centroids for different input graphs should be different. 

\begin{figure*}
   \begin{center}
  \includegraphics[width=\linewidth]{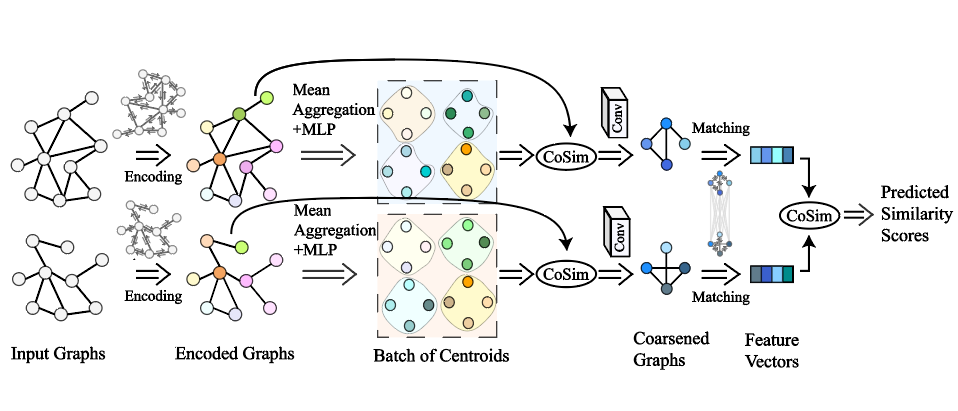}
    \end{center}
 \caption{An overview of our "embedding-coarsening-matching" framework \textsc{CoSim-GNN}. The encoding part aggregates features, respectively, inside the two graphs. The coarsening stage then transforms the encoded graphs into batches of centroids, and these centroids jointly coarsen the graphs. Finally, matching-based feature aggregation is deployed on the coarsened graph pair, and the similarity score is computed based on the two graph-level feature vectors.}
 \label{fig:our model}
\end{figure*}

\section{Proposed Framework} \label{sec: Proposed Framework}
In this section, we respectively introduce our proposed framework \textsc{CoSim-GNN} in details. The overview of \textsc{CoSim-GNN} is shown in Fig. \ref{fig:our model}. 
We use bold font for matrices and tilt font for function. 
The right superscript of a matrix stands for graph number indicator and the right subscript stands for the stage of the matrix. 
For example, $\textbf{X}_{in}^{(1)}$ means the node feature matrix for the first input graph and $\textbf{A}_{encode}^{(2)}$ means the adjacency matrices for the second graph after encoding. 


\subsection{Problem Formulation}
Given a pair of graph inputs ($G^{(1)},G^{(2)}$), the research goal of graph similarity computation in this paper is to produce a similarity score $y\in$ $\mathcal Y$. The graph $G^{(1)}$ is represented as a set of $n^{(1)}_{in}$ nodes with a feature matrix $\mathbf{X}^{(1)}_{in}\in \mathbb{R}^{n^{(1)}_{in}\times d^{(1)}_{in}}$ and an adjacency matrix $\mathbf{A}^{(1)}_{in}\in \mathbb{R}^{n^{(1)}_{in}\times n^{(1)}_{in}}$. Similarly, the graph $G^{(2)}$ is represented as a set of $n^{(2)}_{in}$ nodes with a feature matrix $\mathbf{X}^{(2)}_{in}\in \mathbb{R}^{n^{(2)}_{in}\times d^{(2)}_{in}}$ and an adjacency matrix $\mathbf{A}^{(2)}_{in}\in \mathbb{R}^{n^{(2)}_{in}\times n^{(2)}_{in}}$.

When performing graph similarity prediction tasks, $y$ is the similarity score $y \in (0,1]$; when performing graph pair classification tasks, $y \in \{-1,1\}$ indicates whether the graph pair is similar or not. The model is trained based on a set of training triplets of structured input pairs and scalar output scores ($G_1^{(1)},G_1^{(2)}$,$y_1$), ($G_2^{(1)},G_2^{(2)}$,$y_2$), ...,
($G_N^{(1)},G_N^{(2)}$,$y_N$), where $N$ is the size of training set. 

\subsection{Encoding} \label{subsec: encoding}
In the first stage, given two input node sets denoted as $\mathbf{X}^{(1)}_{in}\in \mathbb{R}^{n^{(1)}_{in}\times d^{(1)}_{in}}$ and $\mathbf{X}^{(2)}_{in}\in \mathbb{R}^{n^{(2)}_{in}\times d^{(2)}_{in}}$, as well as the adjacency matrix $\mathbf{A}^{(1)}_{in}\in \mathbb{R}^{n^{(1)}_{in}\times n^{(1)}_{in}}$ and $\mathbf{A}^{(2)}_{in}\in \mathbb{R}^{n^{(2)}_{in}\times n^{(2)}_{in}}$, 
we employ $k$ encoding layers $(F_{1},F_{2},...F_{k})$ on the two input graphs respectively for embedding feature of nodes and transform the feature dimension of nodes to $d_{encode}$, just as shown in Eq. \eqref{abstracted equation for encoding} below:
\begin{equation}
\label{abstracted equation for encoding}
{\mathbf{X}_{encode}^{(i)}, \mathbf{A}_{encode}^{(i)}}=F_k(F_{k-1}(...F_{1}(\mathbf{X}_{in}^{(i)}, \mathbf{A}_{in}^{(i)})))
\end{equation}

where $i=1,2$, $\mathbf{X}_{encode}^{(i)}\in \mathbb{R}^{n^{(i)}_{in}\times d_{encode}}$, $\mathbf{A}_{encode}^{(i)}\in \mathbb{R}^{n^{(i)}_{in}\times n^{(i)}_{in}}$. Eq. \eqref{description equation for single encoding layer} shows that $F_j\ (j=1,2,...,k)$ is consisted of a graph convolution layer $GC$, a non-linear activation $\sigma$ and a batchnorm layer $bn$.
\begin{equation}
\label{description equation for single encoding layer}
F(\mathbf{X},\mathbf{A})=bn\left(\sigma(GC(\mathbf{X},\mathbf{A}))\right)
\end{equation}
We use ReLU for $\sigma$ here and graph embedding layers $GC$ can be GCN \cite{NIPS2016_6081}, GAT \cite{velivckovic2017graph} or GIN \cite{Xu2018HowPA}.

\subsection{Coarsening} \label{subsec: coarsening}
What this stage does is to pool the encoded graphs, $\mathbf{X}_{encode}^{(i)}$ and $\mathbf{A}_{encode}^{(i)}$, to coarsened graphs, $\mathbf{X}_{pool}^{(i)} \in \mathbb{R}^{n_{pool} \times d_{pool}}$ and $\mathbf{A}_{pool}^{(i)} \in \mathbb{R}^{n_{pool} \times n_{pool}}$. The overall transformation is shown in Eq. \eqref{overall equations of pooling layer}, where $\sigma$ is a non-linear activation function, $\mathbf{W}\in \mathbb{R}^{d_{encode} \times d_{pool}}$ is a trainable parameter matrix standing for a linear transformation and $\mathbf{C}^{(i)}\in \mathbb{R}^{n_{pool}\times n_{in}^{(i)}}$ is an assignment matrix representing a projection from the original node number to pooled node number. As $\mathbf{C}^{(i)}$ assigns weights for nodes in the input graph to nodes in the coarsened graph, it indeed stands for an intra-attention mechanism. Note here that because coarsening is done inside seperate graph, we eliminate $i$ in the following detailed explanations.
\begin{equation}
\label{overall equations of pooling layer}
\begin{split}
  \textbf{X}_{pool}^{(i)} &=\sigma\left( \textbf{C}^{(i)} \textbf{X}_{encode}^{(i)} \textbf{W} \right)   \\[1mm]
  \textbf{A}_{pool}^{(i)} &=\sigma\left(  \textbf{C}^{(i)} \textbf{A}_{encode} (\textbf{C}^{(i)})^{\mathsf{T}} \right) 
\end{split}
\end{equation}
Different approaches are adopted to compute the assignment matrix $\mathbf{C}$ in different pooling layers. To make the pooling layers adaptive and input graph related, we first generate $h$ batches of centroids $\textbf{K}\in \mathbb{R}^{h\times n_{pool}\times d_{encode}}$ based on the encoded graph (permutation invariance remains) and then compute and aggregate the relationship between every batch of centroids and the encoded graph, leading to the final assignment matrix $\mathbf{C}$.
Detailed ablation study about pooling layers is conducted in \hyperref[sec: Experiments and Results]{Section 5}, which demonstrates the effectiveness of this pooling operation.

As shown in Eq. \eqref{generation of memory heads}, an average aggregation $F_{avg}$ over the encoded graph is deployed, transforming $\mathbf{X}_{encode}\in \mathbb{R}^{n_{in}\times d_{encode}}$ to $\mathbf{X}_{avg}\in \mathbb{R}^{1 \times d_{encode}}$. 
Then a multiple layer perceptron (\textsc{MLP}) is applied to map $\mathbf{X}_{avg}$ to $\textbf{K}\in \mathbb{R}^{(h\times n_{pool})\times d_{encode}}$, after which $\mathbf{K}$ is reshaped to $\mathbb{R}^{h\times n_{pool}\times d_{encode}}$. 
Note that the generation of batches of centroids here is dependent on the encoded graph (input of this layer), which makes sense because centroids for different input graphs in the training and testing set should be different. 
Besides, due to the mean aggregation, we keep the permutation invariance property in our proposed pooling method, one of the most important properties for graph-related deep learning architecture.

\begin{equation}
\label{generation of memory heads}
\mathbf{K} = MLP\left(F_{avg}\left( \mathbf{X}_{encode}\right)\right)
\end{equation}
Then we compute the relationship $\textbf{C}_p\in \mathbb{R}^{n_{pool}\times n_{in}}\ (p=1,2,...,h)$ between every batch of centroids $\textbf{K}_p\in \mathbb{R}^{n_{pool}\times d_{encode}}\\ (p=1,2,...,h)$ and $\mathbf{X}_{encode}\in \mathbb{R}^{n_{encode}\times d_{encode}}$. A simple cosine similarity is deployed here and leads to satisfactory pooling performance, as described in Eq. \eqref{dist calculation of pooling layers}, where a row normalization is deployed in the resulting similarity matrix.
\begin{equation}
\label{dist calculation of pooling layers}
\begin{split}
\textbf{C}_p &= \operatorname{cosine} \left(\mathbf{K}_{p}, \mathbf{X}_{encode}\right) \\[1mm]
\textbf{C}_p &= \operatorname{normalize}\left(\textbf{C}_p\right)
\end{split}
\end{equation}
We finally aggregate the information of $h$ relationship $\textbf{C}_p\in \mathbb{R}^{n_{pool}\times n_{in}}$. 
In Eq. \eqref{conv on dist of pooling layers}, we concatenate the centroids of $h$ groups $\textbf{C}_p\ (p=1,2,...,h)$ and a learnable weighted sum $\Gamma_{\phi}$ is deployed on the first dimension, resulting in the assignment matrix $\mathbf{C}$.

\begin{equation}
\label{conv on dist of pooling layers}
\begin{small}
\mathbf{C} = \Gamma_{\phi}\left(\mathop{\lVert}\limits_{p=1}^{|h|} \mathbf{C}_{p}\right)
\end{small}
\end{equation}
\subsection{Matching} \label{subsec: matching}

As shown in Eq. \eqref{overall description of matching model}, this part aims to compute similarity based on the interaction of two coarsened graphs. $f_{match}$ takes the two coarsened graphs as input and generates feature vectors $\mathbf{X}_{final}^{(i)}$ for respective graph. Note that $i$ and $j$ in the first equation in Eq. \eqref{overall description of matching model} indicate two different graphs in the graph pair, that is when $i=1$, $j=2$ and when $i=2$, $j=1$. Then a cosine similarity is deployed for the similarity score. Finally we compute the Mean Squared Loss between the score and the ground truth similarity $GT$ for backward update.
\begin{equation}
\label{overall description of matching model}
\begin{split}
\mathbf{X}_{final}^{(i)}&=f_{match}\left(\mathbf{X}_{pool}^{(i)}, \mathbf{A}_{pool}^{(i)}, \mathbf{X}_{pool}^{(j)},\mathbf{A}_{pool}^{(j)}\right) \\[1mm]
\mathbf{Score} &= cosine\left(\mathbf{X}_{final}^{(1)}, \mathbf{X}_{final}^{(2)}\right)\\[1mm]
\mathbf{Loss} &=\frac{1}{batchsize}\sum\limits_{i=1}^{batchsize}\left(Score(i)- GT(i)\right)^2
\end{split}
\end{equation}
Several matching mechanisms such as \textsc{GSimCNN} \cite{bai2018convolutional} and \textsc{GMN} \cite{li2019graph}, can be deployed for $f_{match}$ and here we give a simple example. As shown in Eq. \eqref{overall equation for GMN}, there are several propagators $f_P$ and one aggregator $f_A$ to transform the pooled graphs to respective feature vectors with cross-graph interactions. Note here that we will eliminate the subscript $pool$ of matrix $X$ and $A$ in the following detailed explanations. 
\begin{equation}
\label{overall equation for GMN}
\mathbf{X}_{final}^{(i)}=f_{A}\left(f_{P}...\left(f_{P}\left(\mathbf{X}_{pool}^{(i)}, \mathbf{A}_{pool}^{(i)}, \mathbf{X}_{pool}^{(j)},\mathbf{A}_{pool}^{(j)}\right)\right)\right)
\end{equation}
In the propagators $f_{P}$, we aggregate both inside features ($\mathbf{I}^{(i)}$) and external features ($\mathbf{M}^{(i)}$). As shown in Eq. \eqref{inside propagate of matching model}, we first propagate features inside respective graph with a \textsc{GAT} \cite{velivckovic2017graph} and add up the features for all the neighbors of every node to form $\mathbf{I}^{(i)}\in \mathbb{R}^{n_{pool}\times d_{pool}}$.
\begin{equation}
\label{inside propagate of matching model}
\begin{split}
\mathbf{X}^{(i)}_{gat} &= \textsc{GAT}\left(X^{(i)},A^{(i)}\right)\\[1mm]
\mathbf{I}^{(i)}(k)&=\sum\limits_{\forall j,\ A^{(i)}(k,m) \neq 0} \mathbf{X}^{(i)}_{gat}(m),\ \ k=1,2,...,n_{pool}
\end{split}
\end{equation}
As for the inter-attention, we compute $\mathbf{M}^{(i)}\in \mathbb{R}^{n_{pool}\times d_{pool}}$ with $f_{cross}$ in Eq. \eqref{crossing propagate of matching model}. First a relationship mask is generated across the graph pair, which involves matrix multiply for the normalized graphs and a softmax activation ($\sigma$). Then the interaction between the two coarsened graphs is realized by applying the mask to the graph $\mathbf{X}^j$ and subtracting $\mathbf{X}^i$ with the masked $\mathbf{X}^j$.
\begin{equation}
\label{crossing propagate of matching model}
\begin{split}
\mathbf{M}^{(i)} &= f_{cross}\left(\textbf{X}^{(i)}, \textbf{X}^{(j)}\right) \\[1mm]
&= \textbf{X}^{(i)} -  \sigma\left(\frac{\mathbf{X}^{(i)}}{\| \mathbf{X}^{(i)}\|_{L2}} \cdot \frac{(\mathbf{X}^{(j)})^{\mathsf{T}}}{\| \mathbf{X}^{(j)}\|_{L2}} \right) \cdot \textbf{X}^{(j)}
\end{split}
\end{equation}
After the above two steps, we apply $f_{node}$ to embrace all these features including the original input by concatenating all these features and deploying an \textsc{MLP} on the concatenated matrix.
\begin{equation}
\begin{split}
\mathbf{X}_{P}^{(i)}&=f_{\text {node }}\left(\mathbf{X}^{(i)},\mathbf{I}^{(i)},\mathbf{M}^{(i)} \right) \\[1mm]
&=MLP\left(\mathop{\lVert}\left(\mathbf{X}^{(i)},\mathbf{I}^{(i)},\mathbf{M}^{(i)}\right)\right)
\end{split}
\label{XIM}
\end{equation}
As for the aggregator, we use the following module proposed in (Li et al., 2015), where $L_{G}$, $L_{gate}$ and $L$ are simply implemented with MLP and the nonlinear activation $\sigma$ is the softmax function.
\begin{equation}
\begin{split}
\mathbf{X}_{final}^{(i)}&=f_{A}\left(\mathbf{X}_P^{(i)}\right) \\[1mm]
&=\operatorname{L}_{G}\left( \sigma \left( \operatorname{L}_{\mathrm{gate}} \left( \mathbf{X}^{(i)}_P \right) \right) \cdot \operatorname{L} \left(  \mathbf{X}^{(i)}_P \right) \right) 
\end{split}
\end{equation}

\setlength{\tabcolsep}{1mm}{
\begin{table*}[h]
\renewcommand\arraystretch{1}
\caption{Time complexity comparison in similarity search problem.}
\centering
\begin{tabular}{@{}l l@{}}
\toprule[1pt]
\textbf{Categories} & \textbf{Time Complexity}  \\ \midrule
Embedding models & $O\left(2\times m+K\times d_{final}\right)$ \\
Matching models & $O\left(\left(2\times m \times 2+n\times n + d_{final}\right)\times K\right)$ \\
CoSim-GNN & $O\left(2\times m+n\times n_{pool}+\left(2\times m_{pool}\times2+n_{pool}\times n_{pool}+d_{final}\right)\times K\right)$\\
 \bottomrule[1pt]
\end{tabular}
\label{sec:Time complexity in similarity search}
\end{table*}}

\section{Time Complexity Analysis in Graph Similarity Search Problem} \label{sec: Time complexity analysis}

For a pair of input graphs $\mathbf{X}^{(1)}_{in}\in \mathbb{R}^{n^{(1)}_{in}\times d^{(1)}_{in}}$ and $\mathbf{X}^{(2)}_{in}\in \mathbb{R}^{n^{(2)}_{in}\times d^{(2)}_{in}}$, we can assume that they separately have $m^{(1)}$, $m^{(2)}$ edges and the final embedded feature vectors for both graphs are $\mathbf{X}^{(1)}_{final}\in \mathbb{R}^{d_{final}}$ and $\mathbf{X}^{(2)}_{final}\in \mathbb{R}^{d_{final}}$. Then we can analyse the time complexity over $n^{(i)}_{in}$, $m^{(i)}$ and $d_{final}$. Note that due to the fact that there exists a lot of variance for each model, we here analyse the simplest cases for each category and the real time consumption will be presented later in the Section \ref{sec: Experiments and Results}. Table \ref{sec:Time complexity in similarity search} shows the time complexity comparison for embedding models, matching models and our proposed framework, which will be discussed in details in the three parts below. In our settings, $n_{pool}$ is far less than $n$, thus our framework costs far less time than matching models do, especially when the node number of the graphs is very large.
\subsection{Embedding models}
Considering the simplest case here for embedding models, we only visit every edge once and deploy two computational operational on the two nodes it connect, which contributes to the feature of local topology. Thus the computation complexity for these cases is $O\left(2\times m^{(i)}\right),\ i=1,2$.
\subsection{Matching models}
Assuming the simplest case here for matching models, we first compute the relationship across $\mathbf{X}^{(1)}_{in}$ and $\mathbf{X}^{(2)}_{in}$. This part involves $n^{(1)}_{in}\times n^{(2)}_{in}$ computational operations because we have to calculate the connection between every node in $\mathbf{X}^{(1)}_{in}$ to all nodes in $\mathbf{X}^{(2)}_{in}$. Then for each input graph, we also visit every edge once and deploy two computational operational on the two nodes it connect, which also makes contribution to the feature of local topology. Thus the computation complexity for these cases is $O\left(2\times m^{(i)}+n^{(1)}_{in}\times n^{(2)}_{in}\right),\ i=1,2$.
\subsection{Our framework}
For our framework, we take the denotation in Section \ref{sec: Proposed Framework}, that is the number of nodes in coarsened graphs is $n_{pool}$. In the embedding stage, we similarly have to visit every edge once and compute twice. Then in the pooling part, we make an transformation from $n^{(i)}_{in}$ dimensional space to $n_{pool}$ dimensional space, which costs us $n^{(i)}_{in}\times n_{pool}$ computational operations. Finally in the matching stage, like what have been analysed in the previous section, we need $2\times m_{pool}^{(i)}+n_{pool}\times n_{pool},\ i=1,2$ operations. Thus the resulting complexity is $O\left(2\times m^{(i)}+n^{(i)}_{in}\times n_{pool}+2\times m_{pool}^{(i)}+n_{pool}\times n_{pool}\right),\\ i=1,2$.

\section{Experiments and Results} \label{sec: Experiments and Results}

\subsection{Datasets} \label{sec: Dataset information}
In the experiments, we use four real-world datasets and four synthetic datasets. 

Note that some existing graph datasets (e.g., AIDS\cite{bai2018simgnn}and LINUX\cite{wang2012efficient}), where the number of nodes is relatively small in each graph, have not been discussed.


\subsubsection{Real-world Datasets} \label{sec: IMDB dataset}

\begin{itemize}[leftmargin=*]
\item \textbf{IMDB} \cite{yanardag2015deep}: IMDB dataset contains more than one thousand graphs indicating appearance of movie actors/actresses in the same movie. We filter the original dataset and choose all the graphs that have 15 or more nodes for experiments. 

\item \textbf{Enzymes} \cite{schomburg2004brenda}: The Enzymes dataset  consists of various types of enzymes and the least node number is 30, which is suitable for our setting.
\item \textbf{COIL} \cite{riesen2008iam}: We follow the same settings as that in \cite{li2019graph}, and treat graphs in the same class as similar.
\item \textbf{OpenSSL} \cite{xu2017neural}: The OpenSSL dataset was built and released for binary code similarity detection with prepared 6 block-level numeric features.
\end{itemize}

We use the first two real-world datasets for similarity regression task and the last two for similarity classification task.

\subsubsection{Synthetic Datasets}
\begin{table}[h]
\renewcommand\arraystretch{1.2}
\caption{Statistics of synthetic datasets.}
\centering
\begin{tabular}[]{|c|cccccc|}
\hline
Dataset &\multicolumn{1}{m{0.95cm}<{\centering}}{Min Nodes} &\multicolumn{1}{m{0.95cm}<{\centering}}{Max Nodes} &\multicolumn{1}{m{0.95cm}<{\centering}}{Avg Nodes} &\multicolumn{1}{m{0.95cm}<{\centering}}{Min Edges} &\multicolumn{1}{m{0.95cm}<{\centering}}{Max Edges} &\multicolumn{1}{m{0.95cm}<{\centering}|}{Avg Edges}
\\ \hline
BA-60  &54 &65 &59.50 &54 &66 &60.06\\
BA-100  &96 &105 &100.01 &96 &107 &100.56\\
BA-200  &192 &205 &199.63 &193 &206 &200.16\\
ER-100 &94 &104 &99.90 &98 &107 &101.43\\
\hline
\end{tabular}
\label{dataset}
\end{table}

To generate a synthetic dataset, we need to generate graphs and give ground truth similarity for graph pairs. A small number of basic graphs is first generated and then we prune the basic graphs to generate derived graphs with two different categories of pruning rules:
BA model \cite{jeong2003measuring} and ER model \cite{erdos1959random}\cite{bollobas2001random}. 
Details about how these two models work will be presented in Appendix \ref{app:BA and ER}. In our experiments, we generate 4 datasets: \textbf{BA-60}, \textbf{BA-100}, \textbf{BA-200} and \textbf{ER-100}, where the first two characters stand for generation rule and behind is the number of nodes in the basic graphs for that dataset. The statistics of datasets are presented in Table \ref{dataset}. These datasets provide new benchmarks to study graph-graph similarity computation.

Creating and using synthetic datasets have two major advantages:
\begin{itemize}[leftmargin=*]
\item  Other than randomly generating graphs with a large number of nodes, the generation process could make the similarity distribution more uniform. Because the random graphs may only have a very small portion of graph pairs are similar, and derived graphs from the same(or different) basic graph can be more similar(or dissimilar) in generation process. Thus, with synthetic datasets, we can obtain more accurate similarity labels (i.e., ground truth similarities).
\item Synthetic datasets allow us to perform more controllable experiments for studying the effect of graph size on different models, which help us better understand strengths and weakness of different models.
\end{itemize}

\subsubsection{Ground Truth Similarity Generation} To calculate the GED ground truth values of graph pairs, we deploy the same mechanism as that in \cite{bai2018simgnn}. Since all the approximate GED algorithms return an upper bound of the exact GED value, they used the minimum GED returned by all approximate GED algorithms for each pair of graph as the ground truth GED. Then with ground truth GED, we can easily get the ground truth similarity. 

For synthetic datasets, we use the minimum value among four evaluation indicators: the three values calculate respectively by \textsc{Hungarian},\textsc{VJ}, and \textsc{Beam} and the GED value we obtain while generating the graph (detailed in Section \ref{sec:5.1.5}). For real-world datasets, we used the minimum value returned by these three algorithms as the ground truth GED. Then we convert this ground truth GED value to the ground truth similarity score with a normalization of the GED followed by a exponential function, resulting in a value among $[0,1]$.
\begin{equation}
\label{similar score calculation}
\begin{split}
    nGED(G_1,G_2)&=\frac{GED(G_1, G_2)}{(|G_1|+|G_2|)/2} \\
    f(x)&=e^{-x}
\end{split}
\end{equation}
where $|G_i|$ represents the total number of nodes in graph $G_i$.
It is worth noting that the neural network approach can learn graph similarity from all the data (similarity value) generated by different traditional algorithmic approaches, thus has the potential to be more effective than any traditional graph similarity computation algorithm. 

\subsubsection{Ground Truth Generation for Synthetic Datasets}
\label{sec:5.1.5}
\begin{figure}[t]
   \begin{center}
   \includegraphics[width=15cm,height=6cm,keepaspectratio]{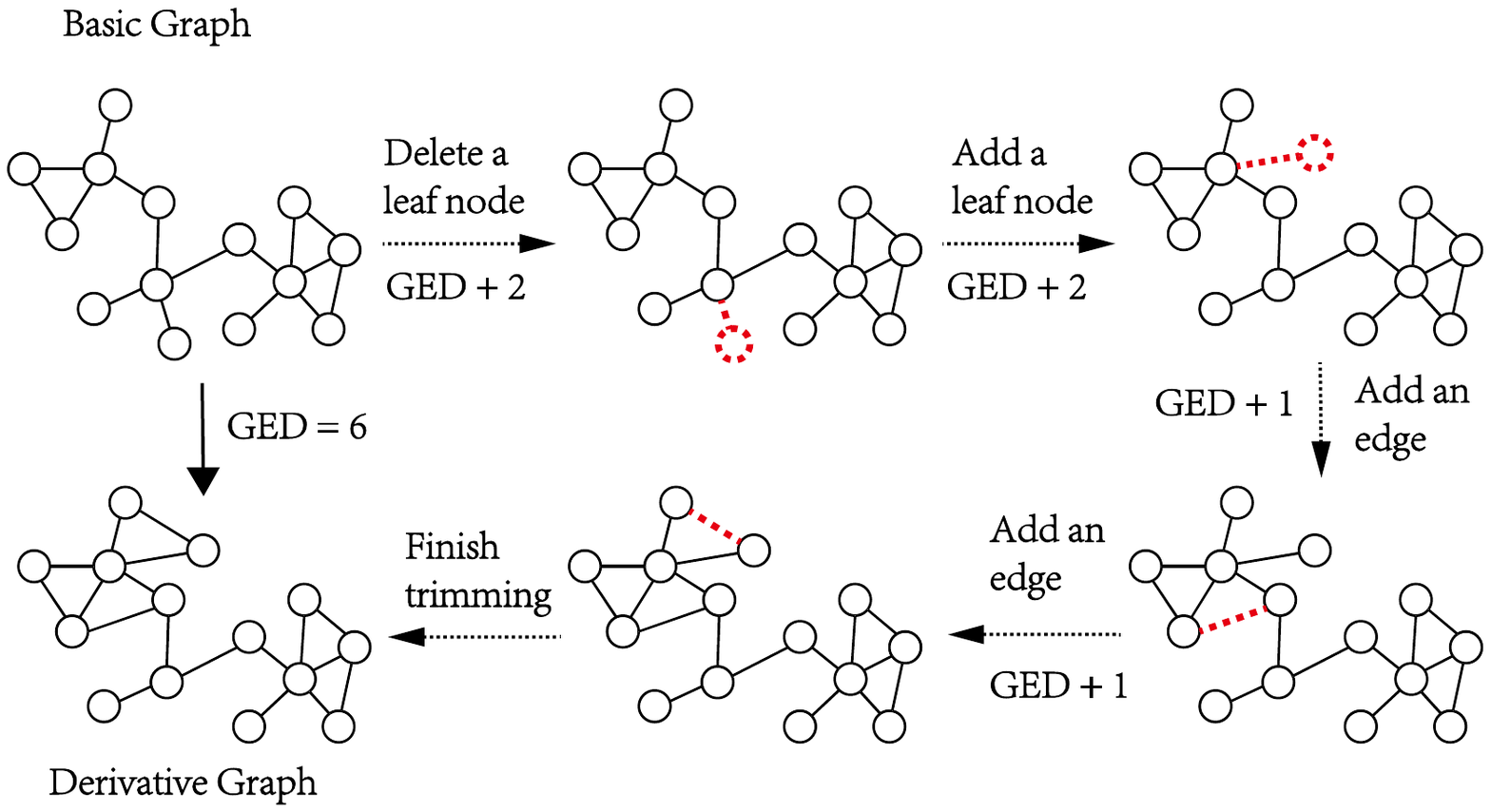} %
    \end{center}
 \caption{Generate a derivative graph with a GED of 6 from the basic graph. It is not necessary to use all three methods in real trimming. Here is only one case.}
 \label{fig:GED}
\end{figure}

Here we mainly discuss details about the GED value obtained when converting basic graphs to derived ones.
As shown in Fig. \ref{fig:GED}, when generating a derivative graph with a specific GED from the basic graph, we randomly select among the above three methods (randomly delete a leaf node, add a leaf node or add an edge) to generate a set of operations, and the sum of GED accumulated by all operations is the exact GED value.
As proposed formerly, we then take the minimum value of the trimming GED and the calculated GED with the three algorithms as the final ground truth GED value.

\setlength{\tabcolsep}{1mm}{
	\begin{table*}[h]
		\renewcommand\arraystretch{1}
		\caption{Results for \textit{MSE} and \textit{MAE} in $10^{-3}$. The three traditional method are involved in the ground truth computation and thus these values are labeled with superscript *.}
		\centering
		\resizebox{\textwidth}{!}{
		\begin{tabular}{@{}c cc cc cc cc cc cc @{}}
			\toprule[1pt]
			\multirow{2}{*}{\textbf{Method}} & \multicolumn{2}{c}{\textbf{BA-60}} & \multicolumn{2}{c}{\textbf{BA-100}} & \multicolumn{2}{c}{\textbf{BA-200}}& \multicolumn{2}{c}{\textbf{ER-100}} & \multicolumn{2}{c}{\textbf{IMDB}} &
			\multicolumn{2}{c}{\textbf{Enzymes}}\\ 
			& MSE        & MAE     &MSE         & MAE         &MSE         & MAE
			&MSE         & MAE &MSE         & MAE
			&MSE         & MAE\\ \midrule
			\textsc{Hungarian}               &186.19$^*$              &332.20$^*$               &205.38$^*$              &343.83$^*$            &259.06$^*$              &379.38$^*$         &236.15$^*$              &421.66$^*$            &2.67$^*$             &16.60$^*$
			&4.67$^*$             &60.70$^*$\\
			\textsc{vj}                    &258.73$^*$              &294.83$^*$             &273.94$^*$             &404.61$^*$      &314.45$^*$             &426.86$^*$         &275.22$^*$              &463.53$^*$ &7.68$^*$             &22.73$^*$ &19.14$^*$             &76.86$^*$  \\
			\textsc{Beam}                    &59.14$^*$             &129.26$^*$       &114.02$^*$
			&206.76$^*$          &186.03$^*$              &287.88$^*$         &104.73$^*$              &226.42$^*$ &0.39$^*$            &3.93$^*$   &$2.97 \times 10^{-5}$ $^*$            &$4.43 \times 10^{-3}$$^*$ \\\midrule
			\textsc{GCN-Mean}                     &   5.85    &  53.92      &  12.53     &90.88    &    23.66  &  127.82     &  16.58    &   92.98 &22.17      & 55.35  & 10.69 & 61.09   \\
			\textsc{GCN-Max}                  &    13.66    & 91.38         &   3.14 & 42.24    &22.77       &  107.58   &   79.09    &   211.07 &47.14     &123.16 & 13.15 & 67.64 \\ \midrule
			\textsc{SimGNN}                  &8.57    &63.80        &6.23             &47.80            &3.06              &32.77       &6.37              &45.30        &7.42            &33.74 & 13.96 & 49.04\\
			\textsc{GSimCNN}                 &5.97     &56.05      &1.86       &30.18       &2.35    &32.64    &2.93     &34.41  &5.01       & 30.43 & 2.48 & 26.79   \\
			\textsc{GMN}             &2.82     &38.38    &4.14       &34.17    &1.16 &26.60 &1.59     &28.68  &3.82      &27.28  & 3.02 & 35.86  \\\midrule
			\textsc{CoSim-CNN}               &2.50  &35.53  &1.49  &27.20    &0.53              &18.44    &2.78  & 33.36 &10.37  &38.03    & 24.64 & 104.05  \\
			\textsc{CoSim-ATT}               &2.04     &33.41     &0.97       &22.95                     &0.73         &16.26       &1.39     &27.27 &\textbf{1.53}       &16.57   & 1.48 & 25.95    \\
			\textsc{CoSim-SAG}               &3.26     &38.85     &3.30     &33.14     &1.91         &35.48      &1.55    &29.84 &1.62     &\textbf{16.08}   & 1.33 & 28.19 \\
			\textsc{CoSim-TopK}               &3.44     &40.87       &1.24      &25.61         &0.88        &20.63     &2.04     &34.28   &1.98     &20.02   & 1.99 & 28.95  \\
			\textsc{CoSim-Mem}                &5.45     &48.07    &1.11       &24.59     &\textbf{0.32}         & \textbf{14.82}    &1.74     &26.78     &1.57       &17.02 & 1.21 & 26.22\\
			\textsc{CoSim-GNN10}                &2.04     &  33.04    &1.01       &23.53              &0.40         &16.43      &1.38     &27.43 &1.68      &17.57 & 1.38 & \textbf{24.73} \\
			\textsc{CoSim-GNN1}               &  \textbf{1.84}    &  \textbf{32.36}      &\textbf{0.95}       &\textbf{22.06}        &0.36         &15.42      &\textbf{1.17}     &\textbf{25.73}    &2.00       &18.62 & \textbf{1.09} & 25.88 \\
			\bottomrule[1pt]
		\end{tabular}}
		\vspace{-1mm}
		\label{Resuls on MSE and MAE.}
    \end{table*}}

\newcolumntype{H}{>{\setbox0=\hbox\bgroup}c<{\egroup}@{}}
\setlength{\tabcolsep}{1mm}{

\begin{table*}[h]
\renewcommand\arraystretch{1.4}
\caption{Results for $\rho$, $\tau$ and p@k}
\centering
\begin{small}
\resizebox{\textwidth}{!}{
\begin{tabular}{c cccc cccc cccc cccc cccc}
\toprule[1pt]
\multirow{2}{*}{\textbf{Method}} & \multicolumn{4}{c}{\textbf{BA-60}} & \multicolumn{4}{c}{\textbf{BA-100}}& \multicolumn{4}{c}{\textbf{BA-200}} &\multicolumn{4}{c}{\textbf{ER-100}} &\multicolumn{4}{c}{\textbf{IMDB}} \\  
                        & $\tau$          & $\rho$          & p@10         & p@20 & $\tau$          & $\rho$        & p@10         & p@20 &$\tau$          & $\rho$         & p@10         & p@20 & $\tau$          & $\rho$        & p@10         & p@20 & $\tau$          & $\rho$         & p@10         & p@20 \\ \hline
hungarian               &0.5772$^*$              &0.7598$^*$         &0.7425$^*$              &0.8438$^*$      &0.6036$^*$              &0.8110$^*$          & 0.6100$^*$             &0.9900$^*$     &0.5810$^*$              &0.7938$^*$          &0.6425$^*$                &0.9400$^*$  &0.3757$^*$                &0.5404$^*$            & 0.6600$^*$               &0.7400$^*$         &0.8318$^*$                &0.9309$^*$           &0.7487$^*$       &0.8092$^*$ \\
vj               &0.0229$^*$    &0.0329$^*$        &0.3500$^*$              &0.5050$^*$        &0.4156$^*$              &0.5837$^*$          &0.4625$^*$              &0.8262$^*$     &0.4310$^*$                &0.6191$^*$          & 0.4850$^*$            &0.8037$^*$    &0.2876$^*$                &0.4030$^*$           &0.5950$^*$                &0.6637$^*$       &0.8334$^*$      &0.9332$^*$           &0.7460$^*$                &0.8151$^*$\\
beam   &0.7434$^*$    & 0.8580$^*$          &0.6775$^*$              &0.9000$^*$       &0.6283$^*$              &0.7867$^*$         &0.6275$^*$              &0.9000$^*$      &0.6521$^*$                &0.7724$^*$            &0.5600$^*$                &0.8350$^*$   &0.4255$^*$                &0.5261$^*$           & 0.5800$^*$               & 0.6937$^*$    &0.9040$^*$      &0.9601$^*$            &0.9092$^*$          &0.9072$^*$\\\hline
GCN-Mean   &0.5329              &0.7564         &0.5800              &0.8688       & 0.5338            & 0.7639         &0.5650              &\textbf{1.0000}      & 0.4946             &0.7347          &0.5000              &0.9500      &0.4708     &0.6704         &0.5175              &0.7862     &0.3620              &0.4664          &0.4868              &0.7026  \\
GCN-Max              &0.5230      & 0.7461      & 0.5450        &0.8662             &0.5304              & 0.7617        & 0.5250             &0.9987     &0.5169              &0.7499          &0.5375              &0.9425      &0.4313              &0.6244      &0.5150              &0.7738       &0.1736              &0.2462      & 0.3908             & 0.4460\\ \hline
SimGNN  &0.5678              &0.7730       &0.7100              &0.8887                &0.5383              &0.7637          &0.5800              &\textbf{1.0000}       &0.4889              &0.7347      &0.5275              &0.9513    &0.5848              &0.7212        &0.6575              &0.8177        &0.3935              &0.5270      &0.5566              &0.6158\\
GSimCNN     &0.6048     & 0.8078          & 0.6775             &0.9050            &0.5932              &0.8079         &0.6450              &\textbf{1.0000}     &0.5761              &0.7958      & 0.6075             &0.9500   &0.6093              &0.8041      & 0.7100             & 0.8488      &0.4987      &0.6626          &0.6210         &0.6414\\
GMN       &0.5601              &0.7753          &0.6150              &0.8800         &0.5499              &0.7755         &0.5675              &\textbf{1.0000}     &0.6034              &0.8140        &0.6375              &0.9500    &0.6538              &0.8363     &0.7650              &\textbf{0.8875}      &0.5538   &0.6956     &0.6566      &0.7187\\\hline
CoSim-CNN                                &0.6059              &0.8057          & 0.7475             &0.9050             &\textbf{0.6335}     & \textbf{0.8332}         &  \textbf{0.7100}            &\textbf{1.0000}  &0.6020             &0.8092      &0.6850             &0.9512      &0.5998   &0.8012          &0.7050              &0.8150     &0.5499              &0.6855      &0.6421              &0.7032\\ 
CoSim-ATT    &0.6076              &0.8083         & 0.6580             &0.8887             &0.6063     & 0.8146         &  0.6550            &\textbf{1.0000}       &\textbf{0.6315}              &\textbf{0.8330}      & 0.6950             &0.9525  &0.6226     & 0.8178     &  0.7800           &0.8025   &0.6051  &0.7237      &0.6632 &0.7343\\ 
CoSim-SAG  &0.6282    &0.8221     &0.6925              &0.8937             &0.6265              &0.8313       &0.6650              &\textbf{1.0000}     &0.5593    &0.7837           &0.5575              &0.9500     &0.5451              &0.7251           &0.6333              &0.8265    &0.6287  &0.7628      &0.6934 &0.7671\\
CoSim-TOPK    &0.6081              &0.8059          &0.6250              &\textbf{0.9175}            &0.5962              &0.8073      &0.5850              &\textbf{1.0000}    &0.5589              &0.7822      &0.5889              &0.9500    &0.5726              &0.7775      &0.7075     &0.8175    &0.5802  &0.7155      &0.6118 &0.6941\\
CoSim-MEM  &0.5529              &0.7704       &0.5625              &0.8837             &0.5763              &0.7962          &0.6275              &\textbf{1.0000}     &0.6184              &0.8252      &\textbf{0.7025}              &0.9512     &0.6481              &0.8308       &\textbf{0.7975}   &0.8100     &0.6057  &0.7486      &0.6434 &0.7520\\
CoSim-GNN10  &\textbf{0.6485}              &\textbf{0.8332}          &\textbf{0.8000}              &0.8775             &0.6261             &0.8315       &0.6475              &\textbf{1.0000}     &0.5950              &0.8097         &0.6875              &0.9500    &\textbf{0.6649}              &\textbf{0.8520}      &0.7775     &0.8500  &0.6629  &0.8050      &0.6868 &0.7697  \\
CoSim-GNN1    &0.6375              &0.8297      &0.7750              &0.9000            &0.6076              &0.8144           &0.6525             &\textbf{1.0000}      &0.6060              &0.8150      &0.6275           &\textbf{0.9550}   &0.6478              &0.8304       &0.7806             &0.8363      &\textbf{0.6989}  &\textbf{0.8277}    &\textbf{0.7276} &\textbf{0.7816} \\\bottomrule[1pt]
\end{tabular}}
\end{small}
\label{BA-60 and BA-100}
\end{table*}
}

\subsection{Experiment Settings}
We randomly divide each dataset to three sub-sets containing 60\%, 20\% and 20\% of all graphs for training, validating and testing (the same as that in \cite{bai2018simgnn}).

\subsubsection{Evaluation Metrics}
We apply six metrics to evaluate all the models: \textit{TIME}, \textit{Mean Squared Error (MSE)}, \textit{Mean Absolute Error (MAE)}, \textit{Spearman's Rank Correlation Coefficient ($\rho$)} \cite{spearman1961proof}, \textit{Kendall's Rank Correlation Coefficient ($\tau$)} \cite{kendall1938new} and \textit{Precision at k (p@k)}. \textit{TIME} is the least average time a model need to compute similarity over one graph pair with the strategy similar to what we discuss in Section \ref{sec: Time complexity analysis}. 

\textit{MSE} and \textit{MAE} measure the average squared/absolute difference between the predicted similarities and the ground-truth similarities. 
\textit{$\tau$} and \textit{$\rho$} measure how well the predicted results match the ground-truth. For \textit{p@k}, we compute the intersection of predicted top \textit{k} similar results and the ground-truth top \textit{k} similar results and divide it by \textit{k}.

\subsubsection{Baseline Methods}
There are three types of baselines. The first category consists of traditional methods for GED computation, where we include \textit{A*-Beamsearch (Beam)} \cite{10.1007/11815921_17}, \textit{Hungarian} \cite{kuhn1955hungarian} \cite{riesen2009approximate}, and \textit{Vj} \cite{jonker1987shortest} \cite{10.1007/978-3-642-20844-7_11}. \textit{Beam} is one of the variants of A* algorithm and its time complexity is sub-exponential. \textit{Hungarian} based on the Hungarian Algorithm for bipartite graph matching, and \textit{Vj} based on the Volgenant and Jonker algorithm, are two algorithms in cubic-time. The second category is made up of embedding models, including \textit{GCN-Mean} and \textit{GCN-Max} \cite{NIPS2016_6081}.
The third category consists of matching models and we here involve \textit{GMN} \cite{li2019graph} and two matching-based models: \textit{SimGNN} \cite{bai2018simgnn} and \textit{GSimCNN} \cite{bai2018convolutional}.

\subsubsection{Settings in Our Proposed Framework}
We provide experiments on seven kinds of variants of our framework to demonstrate its scalability and the effectiveness of the adaptive pooling operation. The names and settings of these seven variants are shown in Table \ref{Model description}. The parameters in our framework are as below: $k=3$, $n_{encode}=64$ (same in baselines), $h=5$, $m=5$, $l_{ins}=2$, $l_{node}=2$, $batchsize=128$ (same in baselines).

We train models for 2000 iterations on BA- and Enzymes datasets, 5000 iterations on IMDB and 10000 iterations on ER-. The model that performs the best on validation sets is selected for testing. 
Experiments are conducted on the same desktop computer with 6 Intel(R) i5-8600K@3.60GHz CPU core. Here, we do not use GPU for acceleration because our experiments show that GPU do not significantly make the learning process faster under our data framework.

\setlength{\tabcolsep}{1mm}{
	\begin{table}[h]
		\renewcommand\arraystretch{1}
		\caption{Settings of different models in our framework.}
		\centering
		\begin{small}
		    \resizebox{\linewidth}{!}{
			\begin{tabular}{@{}l l l l@{}}
				\toprule[1pt]
				\textbf{Model Names} & \textbf{Embedding} & \textbf{Coarsening} & \textbf{Matching}  \\ \midrule
				\textsc{CoSim-ATT} & \textsc{GIN}\ &  \textsc{SimAtt}\ \cite{bai2018simgnn}\ ($n_{pool}=1$) & \textsc{OurMatch}   \\
				\textsc{CoSim-CNN} & \textsc{GIN}\ & \textsc{AdaptivePool}\ ($n_{pool}=10$) & \textsc{GSimCNN}\ \cite{bai2018convolutional}  \\
				\textsc{CoSim-SAG} & \textsc{GIN}\ & \textsc{SAGPool}\ \cite{lee2019self}\ ($n_{pool}=10$) & \textsc{OurMatch}  \\
				\textsc{CoSim-TopK} & \textsc{GIN}\ & \textsc{TopKPool}\ \cite{gao2019graph}\ ($n_{pool}=10$) & \textsc{OurMatch}  \\
				\textsc{CoSim-Mem} & \textsc{GIN}\ & \textsc{MemPool}\ \cite{Khasahmadi2020Memory-Based}\ ($n_{pool}=10$) & \textsc{OurMatch}  \\
				\textsc{CoSim-GNN10} & \textsc{GIN}\ & \textsc{AdaptivePool}\ ($n_{pool}=10$) & \textsc{OurMatch} \\
				\textsc{CoSim-GNN1} & \textsc{GIN}\  & \textsc{AdaptivePool}\ ($n_{pool}=1$) & \textsc{OurMatch}  \\
				\bottomrule[1pt]
			\end{tabular}}
		\end{small}
		\vspace{-3mm}
		\label{Model description}
\end{table}}

\subsection{Results and Analysis}

\subsubsection{Effectiveness and Efficiency}

\begin{figure}[h]
	\begin{center}
		\includegraphics[width=15cm,height=6cm,keepaspectratio]{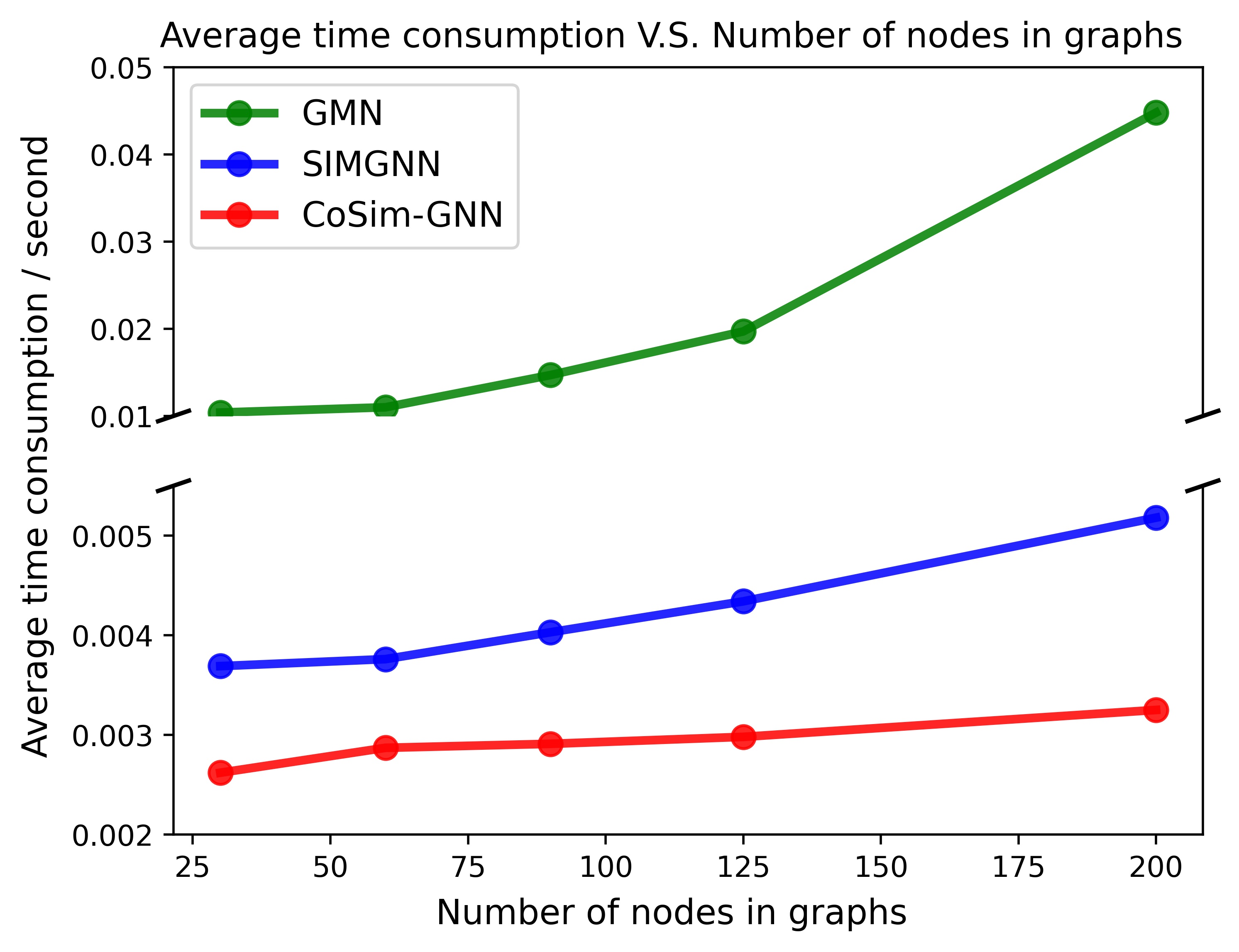} %
	\end{center}
	\caption{Models' running time on different size of graphs.}
	\label{fig: running time}
\end{figure}

Statistic results are shown respectively in Table \ref{Resuls on MSE and MAE.} and \ref{Resuls on Time consumption}. For better understanding the results, we highlight best \textit{MSE} and \textit{MAE} results among all the models in Table \ref{Resuls on MSE and MAE.} and highlight the least time consumption respectively among the four different categories (traditional methods, embedding models, matching models and our proposed models) in Table \ref{Resuls on Time consumption}. It is shown in the statistic that models in our framework outperform matching models in every dataset and our proposed pooling layer achieves the best performance among all the tested pooling method.  Here we emphasize three aspects. Firstly, it can be found that for all the datasets, our framework outperforms all the traditional methods, embedding models and matching models on both \textit{MSE} and \textit{DEV}. Secondly, it is shown that our \textsc{CoSim-GNN1} runs faster than any matching model. When we take comparisons among BA datasets, it is clear that our time consumption keeps low as the node number increases and the gaps on time consumption between our models and matching models keep increasing. Especially when we look at \textsc{CoSim-GNN1} and \textsc{GMN}, our model requires \textbf{1/3}, \textbf{1/5}, \textbf{1/14}, \textbf{1/5}, \textbf{1/5} and \textbf{1/5} as the time consumption of GMN respectively on BA-60, BA-100, BA-200, ER-100, IMDB and Enzymes.

Thirdly, when we look at the results among all the models under our framework, it can be found that in \textbf{4} out of the all 6 datasets, our pooling layer performs better than all other pooling layers, which demonstrates the effectiveness of the adaptive pooling operation.

\setlength{\tabcolsep}{1mm}{
	\begin{table}[h]
		\renewcommand\arraystretch{1}
		\caption{Results for average time consumption on one pair of graphs in milliseconds.}
		\centering
		\begin{small}
		\resizebox{\linewidth}{!}{
			\begin{tabular}{@{}c cc c c c c c c@{}}
				\toprule[1pt]
				\textbf{Method} & \textbf{BA-60} & \textbf{BA-100} & \textbf{BA-200} & \textbf{ER-100} & \textbf{IMDB} & \textbf{ENZYMES}\\ \midrule
				\textsc{Hungarian}               &\textgreater100           &\textgreater100         &\textgreater100        &\textgreater100         &\textgreater100       &\textgreater100   \\
				\textsc{vj}                      &\textgreater100     &\textgreater100         &\textgreater100   &\textgreater100         &\textgreater100   &\textgreater100    \\
				\textsc{Beam}            &\textgreater100          &\textgreater100      &\textgreater100       &\textgreater100      &\textgreater100        &\textgreater100\\\midrule
				\textsc{GCN-Mean}               & \textbf{1.31}         &1.48   &\textbf{1.79}   &1.51               &1.68 & \textbf{1.08}  \\
				\textsc{GCN-Max}                  &1.32        &\textbf{1.46}    &1.85         &\textbf{1.49 }             &\textbf{1.58} &  1.25\\ \midrule
				\textsc{SimGNN}                  &2.92                     & 3.24      & 5.26    &3.46                 &4.34        & 2.62       \\
				\textsc{GSimCNN}                 &\textbf{2.07}   &\textbf{2.73}        &\textbf{5.16}     &\textbf{2.86}         &\textbf{3.16} & \textbf{2.08} \\
				\textsc{GMN}              &5.77      &9.36   &28.57    &10.00          &9.89  & 7.65  \\\midrule
				\textsc{CoSim-CNN}           &1.99   &2.16       &2.67     & 2.19     &2.21     &   1.98 \\
				\textsc{CoSim-ATT}                &1.85         &2.02           & 2.34 &1.99           &2.12  & 1.85       \\
				\textsc{CoSim-SAG}                &2.18       &2.59           &3.97 &2.65         &2.33   & 1.92    \\
				\textsc{CoSim-TopK}               &2.08      &2.28     & 2.51  &2.28       &2.83  & 1.91 \\
				\textsc{CoSim-Mem}                 &3.30       &3.56           & 3.95    &3.50            &3.58 & 3.48           \\
				\textsc{CoSim-GNN10}           &3.29    &3.51            &4.03    &3.63           &3.09   & 3.04    \\
				\textsc{CoSim-GNN1}               &\textbf{1.83}          &\textbf{2.01}            &\textbf{2.31}     &\textbf{1.97}          &\textbf{2.08} &\textbf{1.82} \\
				\bottomrule[1pt]
			\end{tabular}}
		\end{small}
		\vspace{-1mm}
		\label{Resuls on Time consumption}
    \end{table}}
    
We also select graphs with different number of nodes and run \textsc{GMN}, \textsc{SimGNN} and \textsc{CoSim-GNN} for over 100 times. Average time consumption is shown in Fig. \ref{fig: running time}. It can be found that as the average graph size increases, the time consumption of matching models grow much faster than our model's. The reason is that, theoretically, low-level cross-graph interactions in matching models need at least quadratic time complexity over number of nodes of the two graphs, which 
causes them not easily scalable to large graphs.

\subsubsection{Generalization}
Sometimes, due to the limitation of computing resources, we may not be able to train models on very large graph datasets (graphs that have thousands of nodes) efficiently. Therefore, it's interesting and meaningful to explore whether our model can be efficiently trained on small graph dataset and then be effectively applied to infer similarity on large graph pairs.

In this paper, we derive two sub-networks from the academic network of AMiner \footnote{https://aminer.org/aminernetwork}. 
The first sub-network includes 1,397 authors and 1446 edges representing the coauthor relationship and the second sub-network includes 1324 papers and 1977 edges representing the citation relationship. We use these two as the basic graphs, and then use the previously mentioned trimming method (the trimming steps are 10, 20, ..., 100), to get 198 trimming graphs, which with the two basic graphs constitute our Aminer dataset.  Experimental results are in Table \ref{Results on Aminer.}. In our experiments, GMN \cite{li2019graph} can not be deployed to this dataset even with a 32GB RAM (error in memory overflow). It is found that CoSim-GNN outperform the other techniques in both MSE and MAE on the different dataset where graphs have much more nodes. Also the time consumption of CoSim-GNN is even lower than embedding models. This results are very promising in many other application 
including analysis on social network and very large biological molecular.

\begin{table}[h]
\vspace{-3mm}
\renewcommand\arraystretch{1}
\caption{Results on Aminer dataset to show models' generalization ability. MSE and MAE are in $10^{-3}$ and time is in second.}
\centering
\begin{small}
\resizebox{\linewidth}{!}{
\begin{tabular}{@{}ccccc @{}}
\toprule[1pt]
\textbf{Method} & \textsc{GCN-Max} & \textsc{GCN-Mean} & \textsc{gsimcnn} & \textsc{CoSim-GNN} \\ \midrule
MSE                                                   & 7.76                                                                                             &    8.43    & 48.37   & \textbf{3.64}                                           \\ 
MAE                                                        & 59.79                                                                                            &  73.45  & 176.44       & \textbf{50.37}                                            \\ 
Time                                                       & 6.72                                                                                     & 6.96  & 32.00       & \textbf{6.46}                           \\ \bottomrule
\end{tabular}}
\end{small}
\vspace{-3mm}
\label{Results on Aminer.}
\end{table}

\subsection{Study of the Matching Layer}
We conduct an ablation study on BA-100 dataset to validate the key components that contribute to the improvement of our \textsc{CoSim-GNN}.

We focus on equation \eqref{XIM} and select different combination of $\mathbf{X}$, $\mathbf{I}$ and $\mathbf{M}$ in the concatenation operation. Results are shown in Table \ref{Ablation study in matching part}. We can see that  after removing $\mathbf{X}$ or $\mathbf{I}$ or $\mathbf{M}$ from \textsc{CoSim-GNN}, the performance will drop, which proves that all modules are useful for similarity computation. What's more, it can be seen that if $\mathbf{M}$ is removed from the framework, the performance will drop the most. The conclusion here is that all the modules contributed to the framework and the main contribution is brought by $\mathbf{M}$, which stands for cross-graph inter-attention.

\begin{table}[h]
\renewcommand\arraystretch{1}
	\caption{Ablation study in matching part.}
	\centering
	\setlength{\tabcolsep}{0.1mm}{
	\begin{tabular}{p{1cm}p{1.5cm}p{1cm}p{1cm}p{1cm}}
			\toprule[1pt]
    & X\&I\&M & X\&M & M    & X\&I \\\midrule
MSE & \textbf{0.95}    & 1.02 & 1.11 & 2.56 \\
MAE & \textbf{22.06}   &   23.49   &  23.76    &  38.19  \\ \bottomrule
	\label{Ablation study in matching part}
\end{tabular}}
\vspace{-7mm}
\end{table}

\subsection{Graph-graph Similarity Classification}
We conduct similarity classification experiments (classify graph pairs to ``similar'' or ``dissimilar'') on three synthetic datasets we generate and two real benchmark datasets: OpenSSL \cite{xu2017neural} and COIL-DEL (COIL \cite{riesen2008iam}). In Table \ref{classification}, it can be found that ``CoSim'' models achieve the best performance on both synthetic and real datasets in classification task. Experiments also show that more pooling layers lead to better GED regression performance only when the node numbers of graphs are large, i.e. on the BA-200 dataset, and here we only use one coarsening layer for all graph classification tasks.

\begin{table}[h]
	\caption{Accuracy on classification task in $\%$.}
	\centering
	\resizebox{\linewidth}{!}{
	\begin{tabular}{p{2.25cm}p{1.0cm}p{1.1cm}p{1.1cm}p{1.4cm}p{1.0cm}}
			\toprule[1pt]
Method & BA-60                            & BA-100                           & BA-200                           & OpenSSL                         & COIL                            \\ \midrule
GCN-Mean        & 92.56                           & 91.43                           & 91.38                           & 78.93                           & 73.94                           \\
GCN-Max         & 92.19                           & 91.38                           & 90.94                           & 80.72                           & 72.10                           \\
GSimCNN         & 93.78                           & 98.75                           & 98.94                           & 89.50                           & 80.44                           \\
CoSim-Mem       & 93.22                           & 95.00                           & 99.63                           & 90.45                           & 83.85                           \\
CoSim-GNN  & \textbf{97.50} & \textbf{99.38}   & \textbf{99.75}  &  \textbf{94.66}     &  \textbf{88.30} \\     \bottomrule
\end{tabular}}
	\vspace{0.15cm}
	\label{classification}
\end{table}

\subsection{New Benchmarks for Graph-graph Similarity Computation}
In this work, we create a set of synthetic datasets and real-world datasets to study large-scale graph similarity computation. These datasets provide new benchmarks since lacking accurate labels for graph-graph similarity is one of the biggest problem that hinders the development of the field of large-scale graph similarity computation. Detail reasons are as follows.

The first one is the time and memory consumption. It is hard to get graph similarity computation benchmarks mainly because it is quite hard to get the ground truth (labels). To calculate the approximate GED ground truth values of graph pairs, we use the same mechanism as that in \cite{bai2018simgnn}. This mechanism runs in sub-exponential time complexity and has a high memory cost, it is inapplicable to get the ground truths if each graph contains more than hundreds of nodes. Thus graphs in our selected datasets have 20 to 200 nodes and it takes several days to compute the ground truths for each dataset. What’s more, if graphs are too dense, it will also lead to high computational burden when generating labels.

The second one lies in the unbalanced distribution in real-world graph datasets. In many datasets, most of the graph pairs are dissimilar, which result in an uneven similarity distribution after normalization where most of the ground-truth similarities are close to 0. Thus the deep neural network might tend to always output 0 to minimize the loss without actually extracting meaningful representation for graphs.

The third problem is that the approximate GED algorithms are not always accurate especially when the two graphs are large. With synthetic datasets, we can trim and generate derivative graphs while recording the number of trimming steps. We then take the minimum value among this number and the values calculated by the approximation algorithms as the GED with the basic graph, thereby obtaining a more accurate ground-truth similarity. 

\section{Conclusion} \label{sec: Conclusion}
In this paper, we propose \textsc{CoSim-GNN} with an adaptive pooling operation for efficient large-scale graph similarity computation, including three stages: encoding, coarsening and matching. 
Thorough experiments are conducted on various baselines, datasets and evaluation metrics to demonstrate the scalability, effectiveness and efficiency of \textsc{CoSim-GNN}. 

\textsc{CoSim-GNN} opens up the door for learning large-scale graph similarity and the future focus will be on more efficient pooling and matching mechanisms, which are able to further improve the large graph similarity regression and classification in this novel framework. In addition, since obtaining sufficient similarity labels of relatively large-scale graphs is quite hard, it's also promising to explore self-supervised learning methods \cite{liu2020self} for graph-graph similarity approximation.

\section*{Acknowledgement}
This work was supported by the National Natural Science Foundation of China (No.62002035), and the Fundamental Research Funds for the Central Universities (No.2022 CDJQY-022).


\bibliographystyle{ACM-Reference-Format}
\bibliography{sample-base}

\newpage
\vfill
\newpage
\appendix

\section{BA model and ER model's Datasets}
\label{app:BA and ER}
\subsection{Barabási–Albert preferential attachment model.} 
Here we introduce the concept of BA-model, the rules for generating a BA-graph, and how our datasets are produced. 

BA-model \cite{jeong2003measuring} is an algorithm for generating random scale-free networks using a preferential attachment mechanism. 

Specifically, the BA-model begins with an initial connected network of $m_0$ nodes.
New nodes are added to the network one at a time. Each new node is connected to $m\leq m_{0}$ existing nodes with a probability that is proportional to the number of links that the existing nodes already have. 

As a result, the new nodes have a "preference" to attach themselves to the already heavily linked nodes.
 
Our dataset is made up of some basic graphs and derivative graphs that have been trimmed, which solve several challenges:

\begin{itemize}
\item When generating a graph with a large number of nodes randomly, there is a high probability that the generated graphs are dissimilar between each other, which result in an uneven similarity distribution after the normalization described in Eq. \eqref{similar score calculation}.
\item Due to the large number of nodes in each graph, the approximate GED algorithm cannot guarantee that the calculated similarity can fully reflect the similarity of the graph pairs. We trim and generate derivative graphs while recording the number of trimming steps. These steps and the values calculated by the approximation algorithm take the minimum value as the GED with the basic graph, thereby obtaining a more accurate similarity.
\item By trimming different steps, we can generate graphs with different similarities, which is more conducive to the experiment of graph similarity query.
\end{itemize}

There are three types of trimming methods here: delete a leaf node, add an edge, and add a node. Since deleting an edge may have a greater impact on the result of the generated graph, we will not consider this method. We try to trim the base graph without changing the global features of the base graph to generate more similar graph pairs. In this way, we get three datasets according to the following generation rule. 

A BA-graph of $n$ nodes is grown by attaching new nodes each with $m$ edges that are preferentially attached to existing nodes with high degree. We set $n$ to be 60, 100, and 200, respectively, and $m$ is fixed to 1, to generate basic graphs. Each sub-dataset generates two basic graphs, and each base graph is trimmed with different GED distances. For each basic graph, generate 99 trimmed graphs in the range of GED distance 1 to 10. So each sub-dataset consists of two basic graphs and 198 trimmed graphs.

\subsection{Erdős-Rényi graph or a binomial graph model.}
In game theory, Erdős-Rényi model stands for either of two similar models for generating random graphs, named respectively after mathematicians Paul Erdős and Alfréd Rényi. The definition of these two models are as follows:
\begin{itemize}
\item In the $G(n, M)$ model, a graph is chosen uniformly at random from the collection of all graphs which have n nodes and M edges. For example, in the G(3, 2) model, each of the three possible graphs on three vertices and two edges are included with probability 1/3.
\item In the G(n, p) model, a graph is constructed by connecting nodes randomly. Each edge is included in the graph with probability p independent from every other edge. Equivalently, all graphs with n nodes and M edges have equal probability of $p^M(1-p)^{\frac{n}{2}-M}$.
The parameter p in this model can be thought of as a weighting function; as p increases from 0 to 1, the model becomes more and more likely to include graphs with more edges and less and less likely to include graphs with fewer edges. In particular, the case p = 0.5 corresponds to the case where all $2^{\frac{n}{2}}$ graphs on n vertices are chosen with equal probability.
\end{itemize}

A ER-graph connects each pair of n nodes with probability p. Like the BA model, we set n to be 100, and adjust the value of p so that the graph is not too dense. We generate the base graphs and trim them in the same way, so each dataset consists of two basic graphs and 198 pruned graphs too.

\end{document}